# Unraveling Fundamental Properties of Power System Resilience Curves using Unsupervised Machine Learning

Bo Li [a*], Ali Mostafavi [a]

[a] UrbanResilience.AI Lab, Zachry Department of Civil and Environmental Engineering, Texas A&M University, College Station, TX, 77843, USA   Email: libo@tamu.edu

**Abstract:** This study examines hundreds of empirical resilience curves constructed from observational data of recent power outages during extreme weather events to reveal fundamental characteristics of power system resilience. The standard model of infrastructure resilience, the resilience triangle, has been the primary way of characterizing and quantifying resilience in infrastructure systems for more than two decades. However, the theoretical model provides a one-size-fits-all framework for all infrastructure systems and specifies general characteristics of resilience curves (e.g., residual performance and duration of recovery). Little empirical work has been done to delineate infrastructure resilience curve archetypes and their fundamental properties based on observational data. Most of the existing studies examine the characteristics of infrastructure resilience curves based on analytical models constructed upon simulated system performance. There is a dire dearth of empirical studies in the field, which hindered our ability to fully understand and predict resilience characteristics in infrastructure systems. To address this gap, this study examined more than two hundred power-grid resilience curves related to power outages in three major extreme weather events in the United States. Through the use of unsupervised machine learning, we examined different curve archetypes, as well as the fundamental properties of each resilience curve archetype. The results show two primary archetypes for power grid resilience curves, triangular curves, and trapezoidal curves. Triangular curves characterize resilience behavior based on three fundamental properties: 1. critical functionality threshold, 2. critical functionality recovery rate, and 3. recovery pivot point. Trapezoidal archetypes explain resilience curves based on 1. duration of sustained function loss and 2. constant recovery rate. The longer the duration of sustained function loss, the slower the constant rate of recovery. The findings of this study provide novel perspectives enabling better understanding and prediction of resilience performance of power system infrastructure in extreme weather events.

## 1. Introduction

As a component of critical infrastructure systems, the electrical power grid plays a fundamental role in supporting and maintaining the functioning of modern societies. Citizens, industry, and governments heavily rely on electric power to perform economic, social, and administrative activities such as operating machines in factories, providing life support in hospitals, and managing telecommunications and transportation system. Despite its vital role, the power infrastructure system is susceptible to hazard events, including storm, flood, and freezing events. Extreme weather and climate events can cause physical damages to power system facilities or shift the supply-demand relationship to become unbalanced (Choi, Deshmukh, & Hastak, 2016). Thus, large-scale blackouts are one of the most common cascading effects brought by major severe weather events. With the increasing severity and frequency of extreme weather events, power outages pose threats to more people with wider reach as well as higher frequency. The decade from 2011–2021 witnessed roughly 78% more weather-related power outages than did the decade from 2000–2010 (Climate Central, 2022). And Hurricane Ian in 2022 left approximately over 9.62 million people without power; Winter Storm Uri impacted more than 11.7 million people ("Major Power outage events," 2022). Power outages can adversely affect communities' multiple needs for energy services, and further affect other infrastructure and public service provisions considering the interdependency between power systems and other critical infrastructures.

Given the high dependency on and vulnerability of the power system, analyzing and characterizing the resilience of the power system is of utmost importance. Although definitions may slightly vary, resilience of a power infrastructure system in general refers to the system's ability to resist, respond to and recover from disruptions to maintain functionality of delivering energy service to end-users (Bhusal, Abdelmalak, Kamruzzaman, & Benidris, 2020; National Research Council, 2012). To capture and communicate quantitative and qualitative aspects of system behavior during disruptions, the resilience curve is the most often applied tool providing a visual representation of resilience behavior of systems. Resilience curves depict the fluctuation of system performance before, during and after a disruption (Poulin & Kane, 2021). The y-axis of the resilience curve shows the level of performance measure; the x-axis shows the timeline. A typical resilience curve starts from $t_0$, a time when the system stays in normal state. Then the curve records time stamps when disruption occurs and ends, and the performance level during this period. Several key information of resilience can be extracted from resilience curves, such as disruption period and residual performance. Fig 1. shows an example of typical resilience curves.

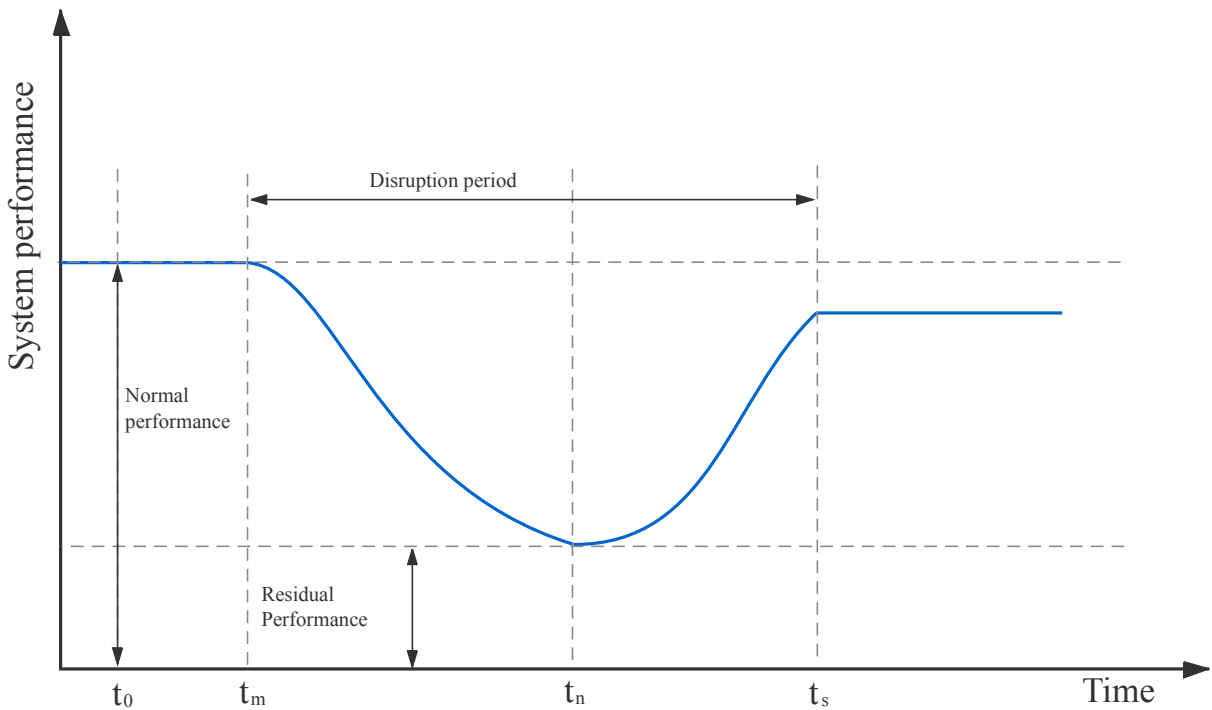

**Fig 1. An example of typical resilience curves**

Despite the widespread application in resilience-related studies, multiple knowledge gaps exist regarding the characteristics of infrastructure resilience curves. First, the resilience triangle has served as the standard paradigm for analyzing resilience behaviors of infrastructure in various contexts for nearly two decades, characterizing the changes of system performance over time as instantaneous performance loss, immediate response, and an approximately linear recovery. This paradigm provides a model to conceptualize resilience behavior for all infrastructure systems, while the one-size-fit-all model fails to capture and represent specific characteristics of resilience curves for each infrastructure type. The follow-up to the resilience triangle paradigm has been the resilience trapezoid paradigm, which recognizes variations in system resilience behaviors, such as cascading failure and pre-recovery degraded performance. However, both the triangular and trapezoidal models are rather conceptual and fail to specify fundamental properties of resilience curves that enable understanding and predicting infrastructure system behaviors under perturbations. Little empirical work to delineate infrastructure resilience curve archetypes and their fundamental properties is based on observational data. Most of the existing studies examine the characteristics of infrastructure resilience curves based on analytical models, which are constructed upon simulated system performance. There is a dire dearth of empirical studies in the field, which hinders our ability to fully characterize, understand, and predict resilience characteristics in infrastructure systems.

To bridge the gap, this study collected power outage data during three major extreme weather events in the United States at fine-grained geographical scales. The power outage data served as a proxy for power system performance loss to delineate resilience curves. Unsupervised machine learning method (i.e., time-series clustering) was adopted to answer the following research questions: (1) What are the main archetypes of the power system resilience curve? (2) What are the fundamental properties of each resilience curve archetype? In the analysis, we empirically revealed two primary archetypes for power system resilience curves, triangular and trapezoidal curves. For each archetype, we identified several fundamental properties that could elucidate the changes of power system performance under perturbations. The findings in this study will advance the understanding of infrastructure resilience behavior in extreme events.

## 2. Literature review

Resilience curves could date back to the work of Bruneau et al. (2003) on seismic community resilience. This study developed a measure $Q(t)$, the service quality of the infrastructure over time, to quantify resilience. The function plot of $Q(t)$ is a resilience curve. Since its birth, the resilience curve has been a powerful tool for researchers to illustrate key information related to disruptive events and affected systems of interest. A typical resilience curve shows the normal performance level, the time when a disruptive event occurred to the system, the disrupted performance of system, the recovery state and performance of stable recovered state (Geng, Liu, & Fang, 2021). Based on the shape of resilience curves, the resilience triangle is the original paradigm widely accepted to describe resilience behavior of infrastructure systems. This paradigm features by instantaneous performance loss after a shock and immediate response for the system performance to bounce back (Poulin & Kane, 2021). Building upon this paradigm, different studies have developed multiple analytical measures to capture features of resilience. Bruneau et al. (2003) computed loss of resilience as the area of resilience triangle. Zobel (2010) considered different resilience loss patterns by illustrating the trade-offs between recovery period length and system initial loss. Bocchini, Frangopol, Ummenhofer, and Zinke (2014) interpreted robustness of the system as the lowest performance level during the disruptive period and rapidity of recovery as the average slope of the recovery path (the hypotenuse of resilience triangle). Cimellaro, Reinhorn, and Bruneau (2010) proposed forms of recovery function as linear recovery, trigonometric recovery, and exponential recovery, and relate the differences to preparedness and resources availability, as well as to societal response.

Despite the ability of resilience triangle model to capture some of the key features of resilience behavior in infrastructure systems, the model provides a rather simplistic and one-size-fits-all characterization. Actually, the performance of an infrastructure system may not drop to its lowest state instantaneously, and the degraded state may last for some time before restoration is initiated. Recognizing this gap, a new pattern called resilience trapezoid was introduced by Panteli, Mancarella, Trakas, Kyriakides, and Hatziargyriou (2017) to overcome the limitations of the resilience triangle model. The resilience trapezoid model portrays all possible phases that an infrastructure system can experience during a disruptive event. Unlike the triangle model, the resilience trapezoid model captures disturbance progress and post-disturbance degradation, as well as the time stamps of the transitions between different stages (Panteli, Trakas, Mancarella, & Hatziargyriou, 2017).

Since the two fundamental models of resilience curves are proposed from a theoretical perspective, there exist two main branches of applying the models: one is to use resilience curves as a conceptual illustration of infrastructure resilience during disruptive events, and the other is to use them in quantitative models and developed metrics to quantify resilience for the infrastructures of interest, usually relying on simulated scenarios (Chen, Liu, Peng, & Yin, 2022). Very few empirical studies have been done to examine the archetypes of resilience curves and their fundamental properties. The dearth of empirical characterization of resilience curves has hindered the understanding and prediction of actual resilience behaviors of infrastructure systems in the real world. To address the gap, this study recorded observational power outage data during extreme weather events as an indicator of power system performance and used them in constructing empirical resilience curves, and then performed unsupervised time-series clustering based on the shape of curves to reveal and characterize archetypes of resilience curve.

## 3. Methodology

This analyzing framework of this paper consists of the following components: (1) collecting power outage reports during extreme weather event and number of energy service customers in all affected geographical units; (2) calculating power outage percentage for each geographical unit at every time point; (3) constructing power system resilience curve for each geographical unit; (4) adopting time series clustering method to identifying power system resilience curve archetypes and key properties. Fig. 2 depicts the overview of the framework.

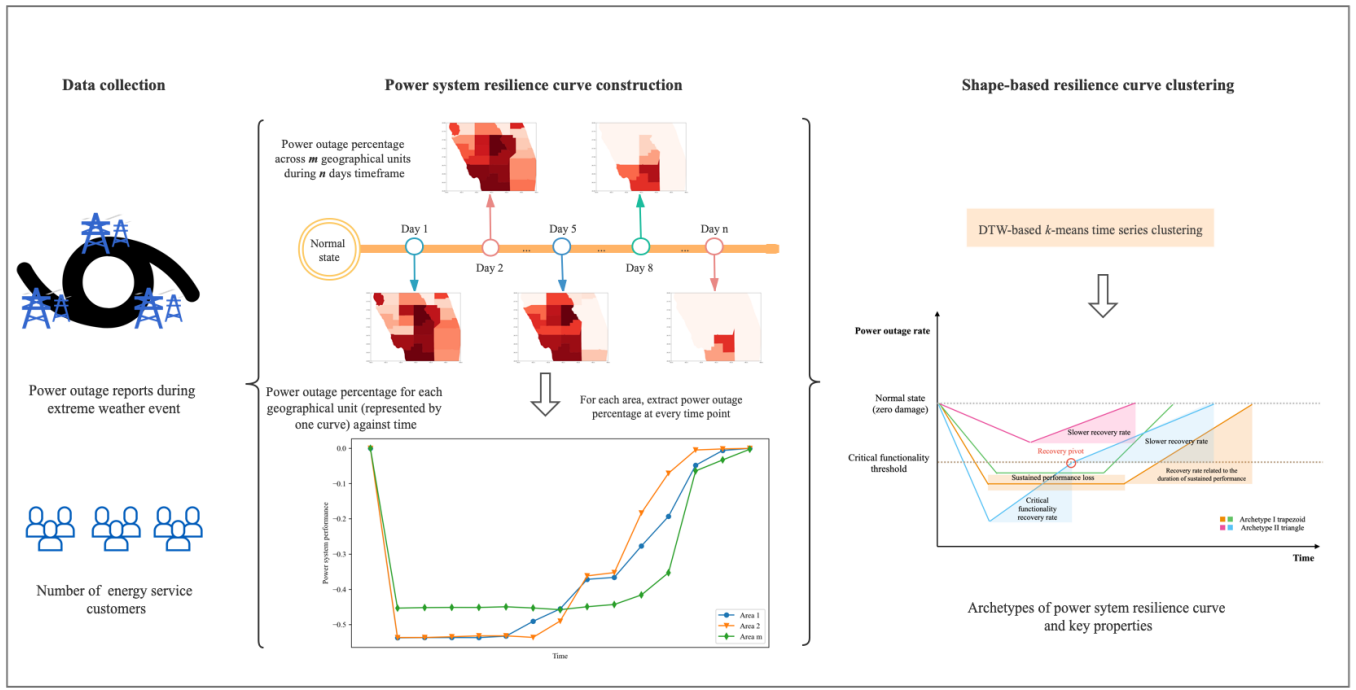

**Fig. 2 Overview of the analysis framework.**

This section first provides brief background information on the hazard events studied in this research and then describes the procedure of constructing and processing raw datasets to build power system resilience curves. Finally, the unsupervised time-series clustering method was introduced and applied to acquire the archetypes of resilience curves.

**3.1 Background information on hazard events**

3.1.1  Ice storm in Austin, Texas (2023)

An ice storm hit the southern part of United States in early 2023, which impacted the Austin area, Texas, from January 30 through February 2. The storm brought several rounds of freezing precipitation, causing widespread ice accumulations. The icing event paralyzed the city for days as the significant ice buildup weighed down trees and powerlines. Fallen trees further toppled powerlines and poles. At the peak of outages, more than 200,000 instances of power outages were recorded across the Austin metro area (NWS Austin/San Antonio, 2023).

3.1.2  Hurricane Irma (2017)

Hurricane Irma made landfall in southern Florida as Category 4 hurricane on September 10, 2017, and then moved into central and northern Florida at Category 3 intensity later (National Weather Service, 2017). The hurricane brought heavy rains and winds, causing widespread power pole damage. Nearly two-thirds of Florida's electricity grid was knocked out by the event, affecting 6.7 million electricity customers. As the count reflects the number of billed accounts, the actual number of affected people could be more since one account can cover more than one person (U.S. Energy Information Administration, 2017).

3.1.3  Hurricane Ida (2021)

Hurricane Ida was one of the most devastating in the recent years, and is listed as the sixth costliest tropical cyclone in United States in the history (NOAA, 2023). As a Category 4 hurricane, Ida made landfall along the southeastern Louisiana coast near Port Fourchon on August 29, 2021, and then affected 20 states in total, including Mississippi, Alabama, Tennessee, Kentucky, Virginia, Maryland (Center, , 2022). Heavy winds downed power lines. A large transmission tower along the Mississippi River west of New Orleans was reported to have collapsed during the event. According to Entergy's estimation, 30,000 utility poles were damaged during Ida, which matches the combining effects of Hurricane Katrina in 2005 and Hurricane Laura in 2020. All of the factors mentioned resulted in massive and severe power outages across eight states, leaving up to 1.2 million electricity customers without service (U.S. Energy Information Administration, 2021).

**3.2 Power outage data collection and preprocessing**

In this study, we used power outage data as an indicator of power system performance. The normal operation of a power system requires functionality of all system components, such as power generation, transmission stations, and powerlines. Thus, power outage data can serve as a comprehensive measure overall power service performance. For Hurricane Ida and the ice storm, this study monitored and collected real-time power outage data during the events. Power outage data for Ida was collected from PowerOutage.US and Entergy power websites. Entergy is the main electricity service provider to the most affected areas in Louisiana, so this study used Entergy outages to reflect the extent of power outages caused by Hurricane Ida. Each record contains Zip code ID, number of affected customers, and last updated time. This study covers period from August 29 through September 11, 2021. For each Zip code region, data points within the same day were averaged to obtain daily values, which could help smooth the data and mitigate the effects of outliers.

In a similar way, real-time power outage data for the ice storm was collected from Austin Energy. The website updates outage information every 10 minutes at Zip code level. We collected the number of affected customers for Zip code regions in Austin at 2-hour intervals from February 3, 2023, through February 9, 2023. The data then were averaged at 6-hour interval, to keep tracking the changes of power outages, while make the data more smoothed.

Power outage data for Hurricane Irma were obtained from Florida Today (Florida Today, 2017), a digital newspaper. The data provides percentage of power outages across Florida at county level. The original data was not collected at a regular temporal interval; instead, one or two representative data points within the same day were provided. For example, power outage percentages records were provided for 6 a.m. and 9 p.m. on September 10, 2017, while on September 11, 2017, only one data point at 12 p.m. was available. To resolve the inconsistency of time intervals, daily averages were computed for days with multiple records available. For days with only one record, single data points were used as a proxy of the daily values. Fig 3. displays the extent of power outage across impacted areas.

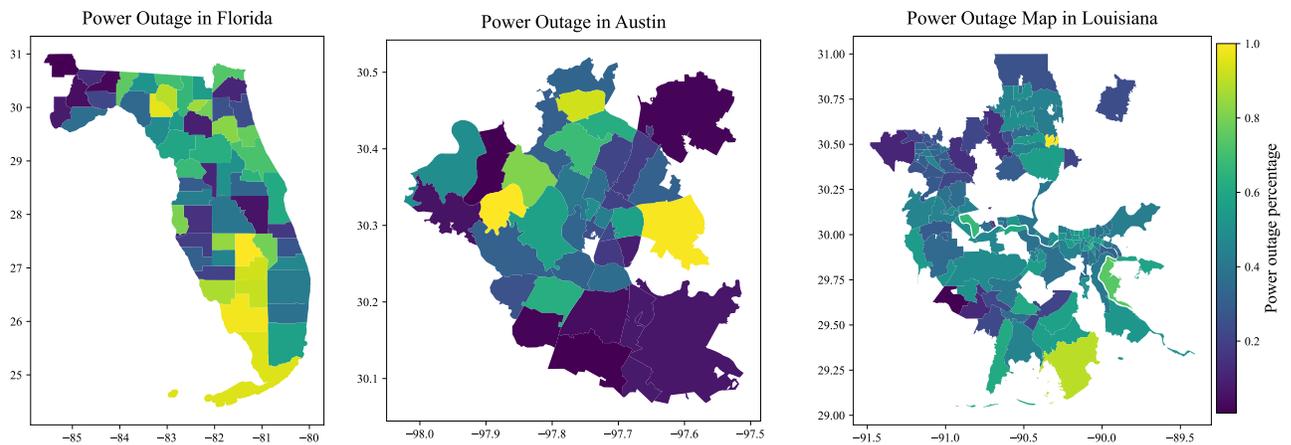

**Fig. 3 Power outage extent across the affected areas**

### 3.3 Power system resilience curves

As discussed in the literature review section, infrastructure resilience curves capture system performance over time during a disruption event. When constructing resilience curves, normalizing performance measures is a common practice to enable comparison across systems and scenarios (Poulin & Kane, 2021). In this study, we used power outage percentage as the normalized system performance. Theoretically, performance can range from 0% to 100%, where 0% designates no service is available, and 100% means no degradation in energy service.

The percentage of outages at each geographical unit was determined by the ratio of the number of affected population and the total population of the geographic unit. Due to the variances of total population size, the absolute number of affected customers may not necessarily the extent of power outage. Thus, we adopted power outage percentage as an indicator. It is important to note that we used the total population of each unit rather than the total number of customers as the nominal values, due to data unavailability. Since each Zip code may have multiple electricity providers, the number of customers of a certain provider in a Zip code may deviate from the number of total population. To alleviate the inaccuracy brought by the deviation, we selected only Zip codes where Entergy and Austin Energy served as the main electricity provider, so that the number of total population can serve as an approximate of number of customers.

For the three hazard events, we constructed three separate datasets, considering differences of scenario, time span, and geographical units. Spatial units with more than 50% missing values were removed, so that the potential inaccuracy incurred by filling missing values could be controlled within an acceptable range. The remaining missing values were filled by linear interpolation. Moreover, spatial units with less than 10% percent of outage during the disruption period were excluded because the researchers were not able to confidently determine whether the outage is due to hazard events or occasional factors. After data cleaning, there remained 222 spatial units in total, 126 Zip codes in Hurricane Ida, 37 Zip codes in the ice storm, and 59 counties in Hurricane Irma. To facilitate visualizing the resilience curves, we set a baseline of as 0% right before the disruption occurred, since power outage percentage is roughly zero when the system operates normally despite that minor variations that may apply. Moreover, all the values were transformed into their respective negative counterparts by employing a unary negation operation, so that the plots were inversed to fit the common shape of resilience curves. We use the 222 empirical resilience curves to specify the archetypes and their fundamental properties.

**3.4 Time-series clustering**

Although power system resilience curves have been established, it is challenging to identify patterns directly from hundreds of empirical curves. Thus, an unsupervised machine learning method was applied to cluster the curves. Every curve actually represents a time series, which is a sequence of observations (power outage percentage in this paper) over the disruption period. This study employed a shape-based time series clustering approach, which partitioned the time-series dataset into a certain number of clusters according to shape similarity. Shape-based clustering usually applies conventional clustering algorithms but modifies similarity measures to make it compatible with time series data (Aghabozorgi, Shirkhorshidi, & Wah, 2015).

This study adopted *k*-means to identify archetypes of resilience curves with similar shape. *K*-means is an unsupervised partition-based clustering algorithm that splits samples into K groups and minimizes the sum of mean squared distance within each group (Kanungo et al., 2002). Given a dataset on *n* time series $T = \{t_1, t_2, \dots, t_n\}$, partitioning $T$ into a total of $k$ clusters, i.e., $C = \{C_1, C_2, \dots, C_k\}$ can be solved by minimizing the objective function J, expressed as:

$$\sum_{j=1}^{k}\sum_{i=1}^{n} D(t_i^j, C_j) \qquad \text{Eq. (1)}$$

where $t_i^j$ denotes the time series $t_i$ in category *j*, and $D(t_i^j, C_j)$ denotes the similarity measurement of the distance between $t_i^j$ and the cluster center of $C_j$).

From Eq. (1) one can see that distance measurement is the most critical option when performing *K*-means. Euclidean distance is the most commonly used metric, yet it is not appropriate to time series. Euclidean distance computes the direct distance between corresponding points while not considering the time dimension of the data. As a result, if two time series are highly similar in their shapes but misalign in time points, Euclidean distance would erroneously measure the two time series as far apart, leading to the failure to capture the shape similarity. To overcome the limitation, this study adopted dynamic time warping (DTW), another distance measurement specific to temporal data. DTW distance is calculated as the square root of the sum of squared distances between each element in X and its nearest point in Y. For better understanding, we illustrate the calculation of DTW distance with an example involving power outage percentages of two Zip codes, namely $Z_p$ and $Z_q$, during a disruptive event. Suppose the two time series for $Z_p$ and $Z_q$ are denoted as $p = (p_1, p_2, \ldots, p_i)$ and $q = (q_1, q_2, \ldots, q_j)$. To determine DTW distance between $P$ and $Q$, a distance matrix $D$ with $i \times j$ elements is first established. The element of the matrix, $d_{ij}$ is calculated as follows:

$$d_{ij} = \sqrt{(p_i - q_j)^2} \qquad \text{Eq. (2)}$$

where $p_i$ is the outage percentage at the $i^{th}$ time point in time series $p$ and $q_j$ is the outage percentage at the $j^{th}$ time point in time series $q$. In the distance matrix $D$, a series of neighboring elements that connect the lower left corner through upper right corner and achieve a minimum of cumulated $d_{ij}$, can be found as a warping path. A warping path can be denoted as $\pi = [\pi_0, \ldots, \pi_k]$, where $\max(m, n) \leq k \leq m + n - 1$, and $\pi_k$ is an index pair $(i_k, j_k)$. DTW distance is the length of the warping path, formulated as

$$DTW(p, q) = min_\pi \sqrt{\sum_{(i,j) \in \pi} d_{ij}} \qquad \text{Eq. (3)}$$

with the path subjecting to boundary conditions, continuity condition and monotonicity condition, about which more information can be found in (Rakthanmanon et al., 2013).

Before performing DTW-based k-means clustering, the number of clusters (i.e., k) needs to be predefined. As the ground truth is not available, the common practice of determining k is to repeatedly execute the clustering algorithm with several k values to identify the optimal. This study provides a more comprehensive approach by combining two performance metrics, namely the silhouette score and the distortion.

The silhouette score provides a measure of how well the data points within each cluster are separated from other clusters. Given a clustering result with *k* clusters, for a resilience curve *i*, let $a(i)$ be the average DTW distance between *i* and all the other resilience curves in the same cluster, and let b(i) be average DTW distance between *i* and all the resilience curves in the nearest cluster, the silhouette score $s(i)$ of resilience curve *i* is computed by:

$$s(i) = \frac{b\{i\}-a\{i\}}{\min\{a(i),b(i)\}} \qquad \text{Eq. (4)}$$

The silhouette score of the entire clustering is given by taking the average $s(i)$ of all resilience curves. The range of the silhouette score is from -1 to, and higher score indicates higher level of separation between clusters.

Similar to the process of employing the silhouette score to determine *k*, the elbow method first chooses a range of possible cluster numbers and runs a clustering algorithm for each *k*. For each round of the clustering, it calculates the within-cluster sum of squared distances from each data point to its assigned cluster center. The value is usually referred as distortion. Given the cluster centers as $c_i$, where *i* ranges from 1 to *k*, and given $S_i$ as the set of time series data points (resilience curves in this study), the distortion can be calculated as

$$distortion = \sum_i DTW(S_i, c_i)^2 \qquad \text{Eq. (5)}$$

where $DTW(S_i, c_i)^2$ denotes the DTW distance between all the resilience curves in cluster $i$ and cluster centers $c_i$. The level of distortion reflects how compactly data points are grouped around their cluster centers, and lower distortion values indicate more compact clusters.

Since the silhouette score and distortion underscore the different aspects of clustering performance, it is a better idea to combine the two measures to find the optimal cluster number, so as to ensure both well separation between clusters and compactness within clusters.

This study used tslearn (Tavenard et al., 2020), a Python machine learning package specific to time-series data to perform tasks, including $k$-means clustering, silhouette score, and distortion calculation.

## 4. Results

4.1 Implementation of DTW-based $k$-means method

Power system resilience curves during the three hazardous events were clustered respectively. For each of the events, a few trials were performed to determine the most appropriate number of clusters ($k$). The initial range of $k$ was set from 2 to 10. The silhouette score and distortion based on DTW distance were computed and plotted as Fig 4. For comparison purposes, similar experiments were performed using Euclidean distance-based $k$-means method. Generally, the results for all the three events show higher silhouette score and lower distortion based on DTW distance than that of Euclidean distance, indicating DTW-based $k$-means method have the greater potential to improve clustering performance. For Hurricane Ida, the highest silhouette appears when $k = 2$, and the second highest peak level is achieved at $k = 3$ and $k = 5$. The distortion plot in Fig. 4b shows that the "elbow point" is at $k = 4$, where the distortion decrease rate gets flatter. As a result of trade-off between the two metrics, the number of clusters was set to 5 so that the silhouette maintains a relatively high level while the distortion is low. Silhouette results for the ice storm shows similar pattern that the peak is reached at $k = 2$ with the second highest level is achieve at $k = 3$ and $k = 7$. However, no significant "elbow point" can be identified from Fig. 4d. Considering the small number of data points collected in this event, large number of clusters can lead to a series of problems (for example $k = 7$), such as overfitting, unstableness of clustering, and lack of representativeness. Thus, $k = 2$ is selected, which also yields the highest silhouette score. For Hurricane Irma, both Fig. 4e and Fig. 4f shows consistent results that $k = 3$ is the "elbow point" and achieves the highest silhouette score. Basically, the decisions of cluster number in this study were the result of comprehensive factors, including the two performance metrics, as well as the number of data points, and cluster interpretability. The final results of clustering for the three natural hazard events are listed in Table 1.

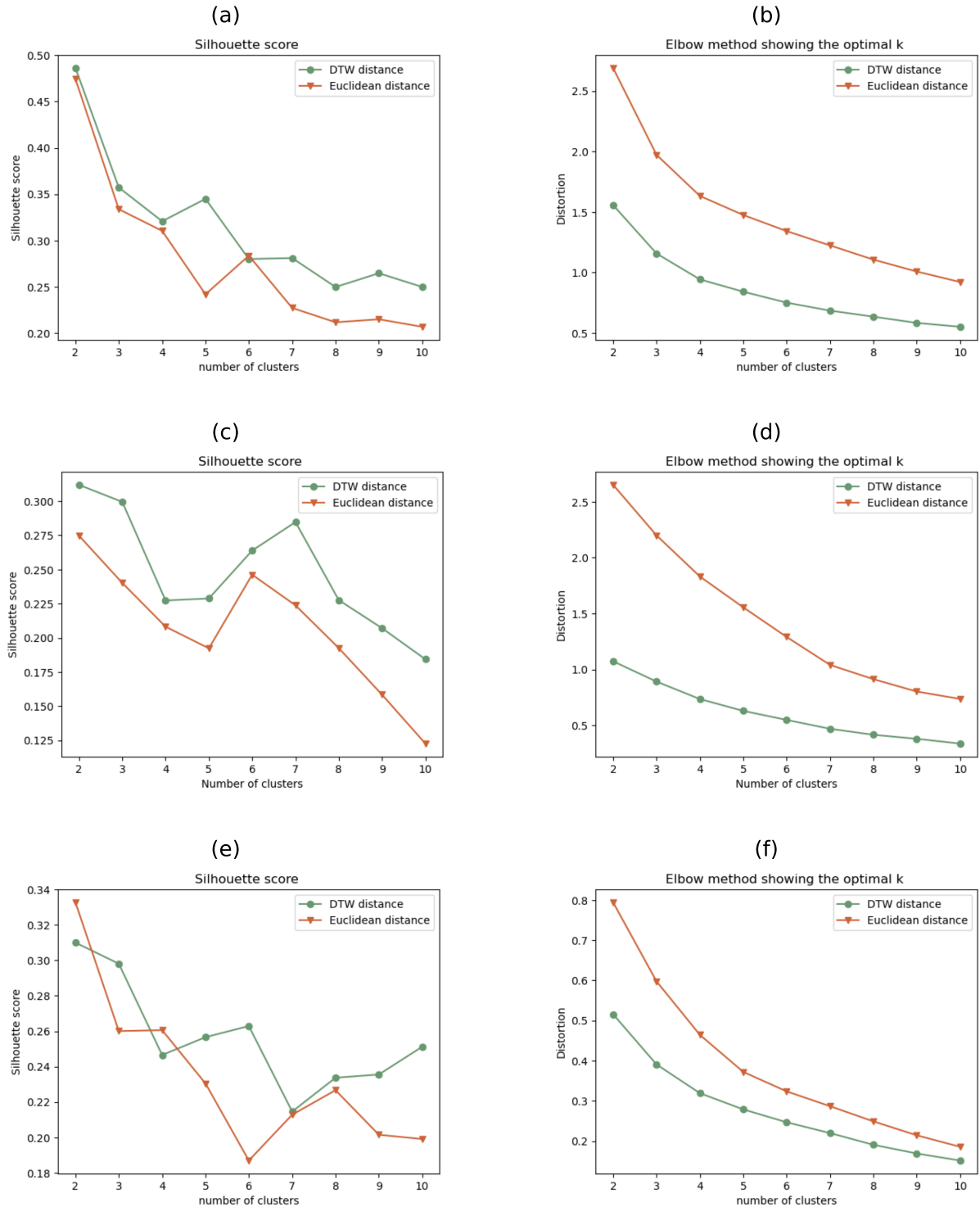

**Fig. 4 The silhouette score and distortion of varying numbers of clusters for the three natural hazard events. a and b.** Silhouette score and distortion for Hurricane Ida; **c and d.** Silhouette score and distortion for the ice storm; **e and f.** Silhouette score and distortion for Hurricane Irma.

Table 1. Summary of clustering result for the three hazard events

| Event | Number of resilience curves | Number of clusters | Silhouette score |
|---|---|---|---|
| Hurricane Ida | 126 | 5 | 0.345 |
| Ice storm | 37 | 2 | 0.312 |
| Hurricane Irma | 59 | 3 | 0.298 |

4.2 Clustering results

The clustering results are displayed in Fig. 5. The plots on the figure show the average curves for each cluster, which represents the shape of the members in each cluster. Fig. 5a displays five clusters, while cluster 3 contains only four Zip codes, which did not provide sufficient confidence on its reliability of showing a specific pattern or just outliers. Thus, out of prudence, cluster 3 in the Hurricane Ida power outage was excluded from the following analysis. The remaining clusters indicated two main archetypes regarding the shape of the curves: triangle and trapezoid. Cluster 1 in Hurricane Ida, both clusters (cluster 1 and 2) in the ice storm and cluster 2 and cluster 3 in Hurricane Irma show a triangular archetype, while other clusters show the trapezoidal archetype. These findings show consistency with the prior research that both triangular and trapezoidal archetypes exist regarding the shapes of resilience curves. Furthermore, the result shows that triangular and trapezoidal patterns could exist in the same event (Hurricane Ida and Hurricane Irma in this study), which, to the best of the authors' knowledge, has not been reported before. Even with the similar extent of performance loss, the power system could still take on different resilience behavior (cluster 1 and 3 in Hurricane Irma) by starting to recover with no delays or remaining in sustained performance loss for some time then starting to recover.

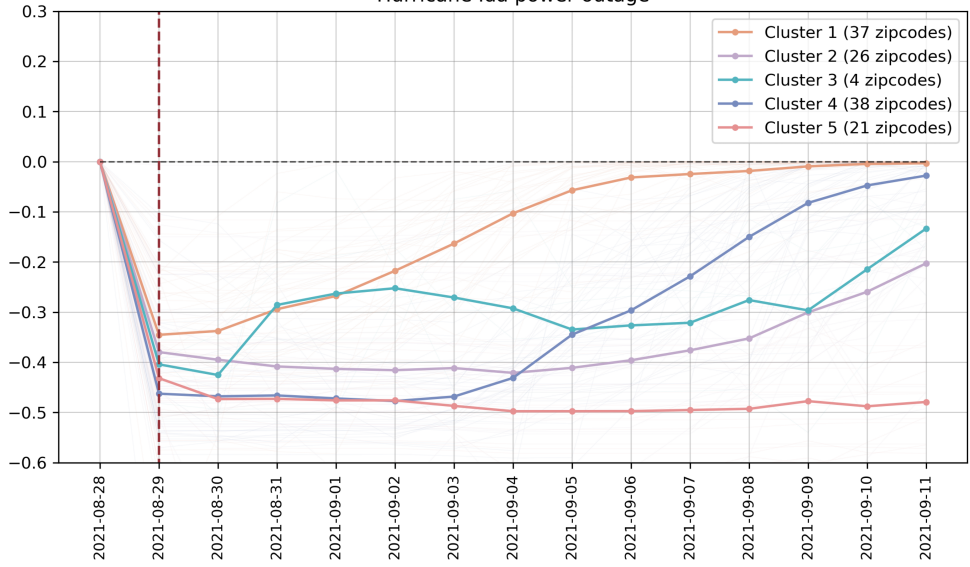

(a) Hurricane Ida power outage

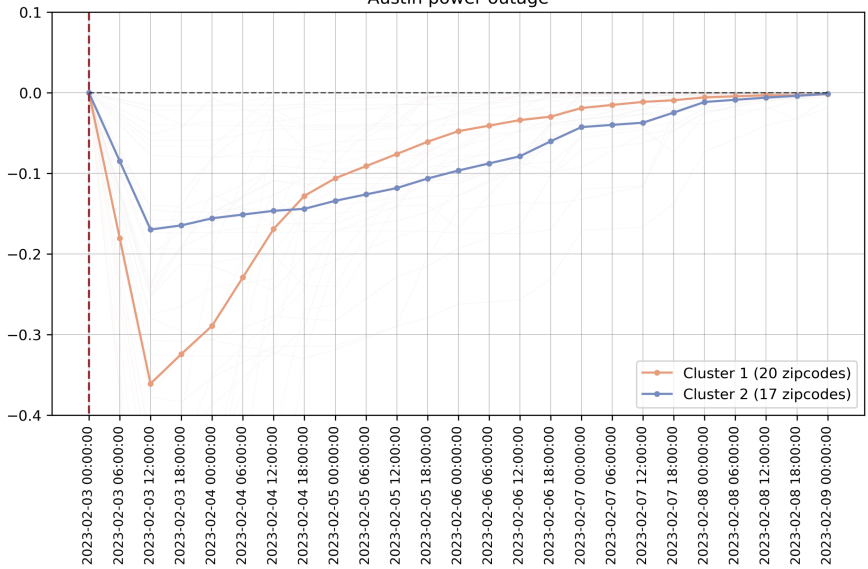

(b) Austin power outage

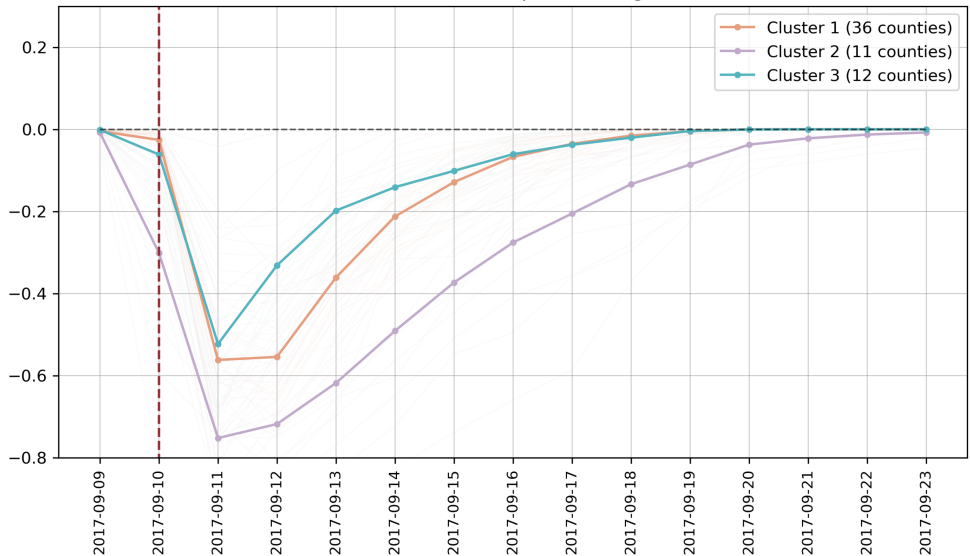

(c) Hurricane Irma power outage

**Fig 5. Clustering result for the three natural hazard events. a.** Five clusters are obtained for Hurricane Ida; **b.** Two clusters are obtained for the ice storm; **c.** Three clusters are obtained for Hurricane Irma.

After specifying the two archetypes of power system resilience curves, this study further investigated the key properties characterizing each archetype. First, we inferred the properties from direct observation on the averaged curves, and then provided supporting evidence based on additional calculations. For trapezoidal curves, this study divided each trapezoidal curve into two stages after the initial performance loss: the curve maintained maximum performance loss level for several days, during which we named as *sustained performance loss stage*, and then started to bounce back with a certain recovery rate. Furthermore, we could speculate upon the relationship between duration of sustained performance loss and recovery rate: the longer the system stays in maximum performance loss state, the slower the recovery would be.

Compared with trapezoidal curves, triangular curves are characterized to have instantaneous recovery, while the recovery processes occur through a two-stage pattern. The system performance first experiences a faster recovery until a certain performance level, and then the recovery rate slows down until the system is fully recovered. Based on the observation, we defined the turning point of recovery rate as *recovery pivot point*, at which the recovery rate changes. The corresponding performance levels for recovery pivot among the triangular curves are quite close, indicating that the performance level of recovery pivot might be an important threshold. We call the performance level the *critical functionality threshold*, since we infer that the power system would recover rapidly to the threshold to restore the critical functionality of the system. The recovery rate after restoring the performance to the critical functionality threshold would be slower. Accordingly, the recovery rate before achieving critical functionality is named critical functionality recovery rate. Critical functionality threshold could also explain the triangular curves with single recovery rate: since the performance loss of such curve does not exceed the threshold, the system follows one recovery rate. Fig. 6 displays both archetypes and associated properties identified from clustering results.

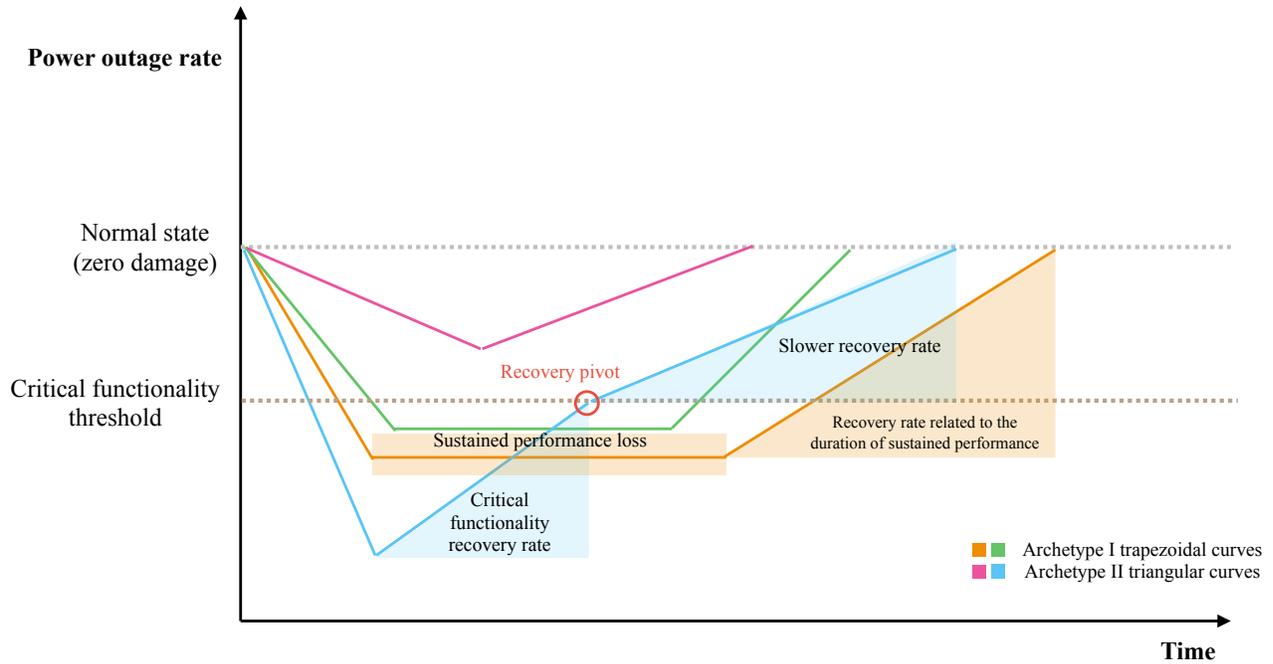

**Fig. 6 Conceptual illustration of archetypes and key properties of power system resilience curve under extreme weather.**

To provide quantitative evidence to support the observations made on the properties of each resilience curve archetype, this study calculated gradients of each data point on the averaged resilience curves and further calculated percentage change of the gradients. Gradients indicate the failure and recovery rate at each time point, while percentage changes indicate the speed for failure or recovery rate to change. Fig. 7 summarizes the archetypes and associated properties of power system resilience, as well as corresponding indicators used to provide quantitative support.

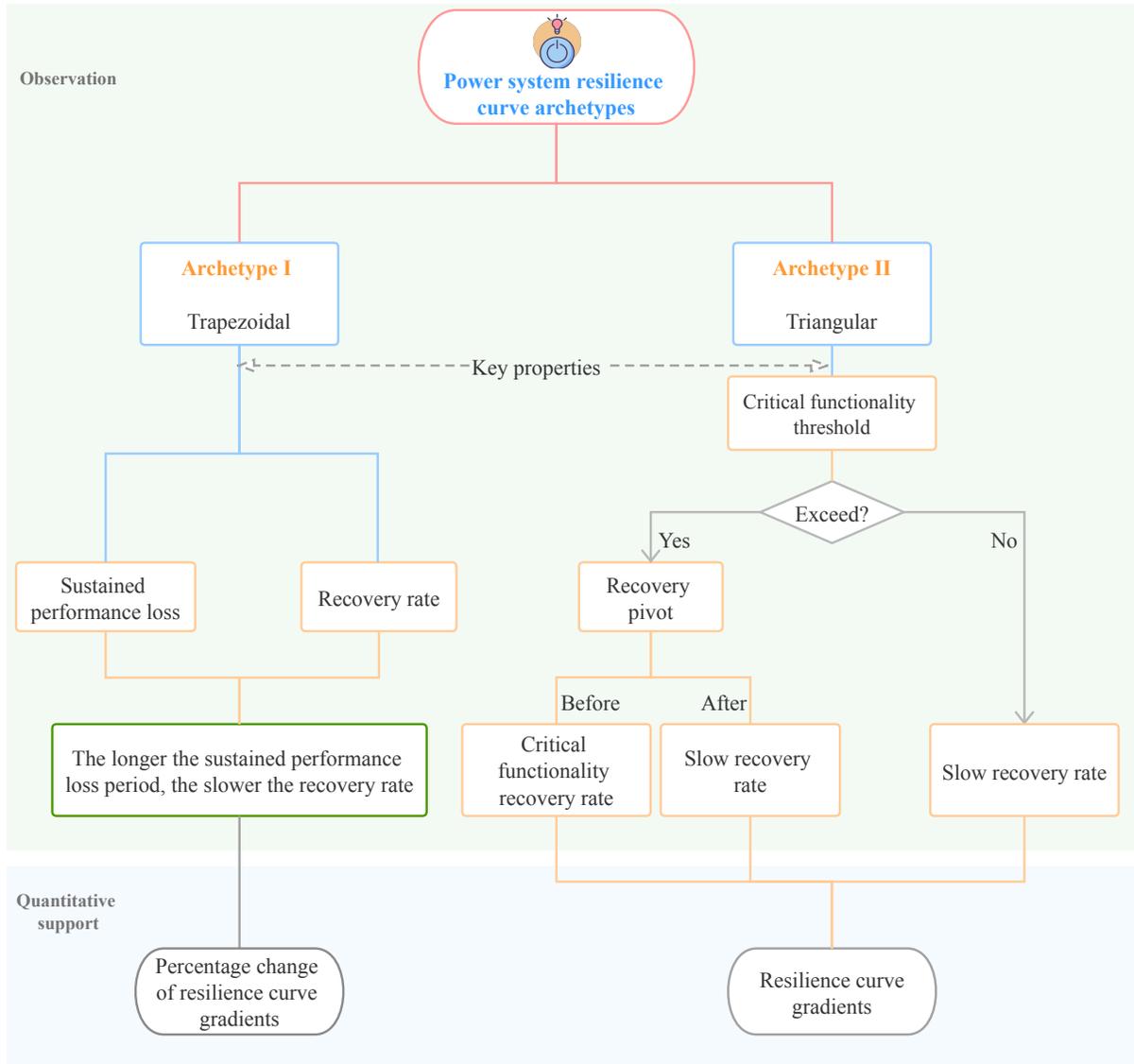

**Fig. 7 Understanding resilience behavior of power systems based on the identified archetypes and their properties.**

The quantitative gradient change analysis results of trapezoidal curves are shown in Fig. 8. Significant peaks can be identified in the plots of gradient percentage change indicating the time point when dramatic changes occurred to the curves, namely the moment when trapezoidal resilience curves start to recover from sustained maximum performance loss. For example, a peak of gradient change stands out on September 5, 2021, for cluster 2 in Hurricane Ida (Fig. 8a), and at the same time, six days of sustained performance loss stage came to an end and the power outage percentage started to decline. Similar patterns also can be found on cluster 4 of Hurricane Ida (Fig 8b). The peak occurred on September 3, 2021, the same day on which power system performance starts to recover. The peaks of the two clusters indicate drastic gradient change of the curves, which are cohesive with the observation of averaged resilience curves. Thus, the corresponding points would be the turning points which separate sustained performance loss stage and recovery stage. Based on that, we calculated the duration of sustained performance loss as the number of days between the day when lowest performance level and the day when the turning point occurred. After the turning point, the performance of the system starts to recover at a constant pace, since no significant peaks of gradient performance changes are observed anymore. From the calculation, we could further infer the relationship between duration of sustained performance loss and recovery rate: the longer the system stays in maximum performance loss state, the slower the recovery would be.

Note that there is a peak in cluster 5 of Hurricane Ida while significant changes were observed neither from the plot of averaged resilience curve nor the plot of gradients. The presence of the peak is due to the plot of cluster 5 being almost horizontal, and the resulting gradients are so small in magnitude that gradient percentage changes of this curve are quite sensitive. Although the peak shows a great change of gradients between data points on September 4 and 5, 2021, the actual gradients of the two points are tiny (0.00531 and 0.000118), which could be neglected. Thus, we ignored the "fake peak" and considered this resilience curve as nearly horizontal. Cluster 5 is an extreme case, with the longest period of sustained loss and zero recovery during the research period.

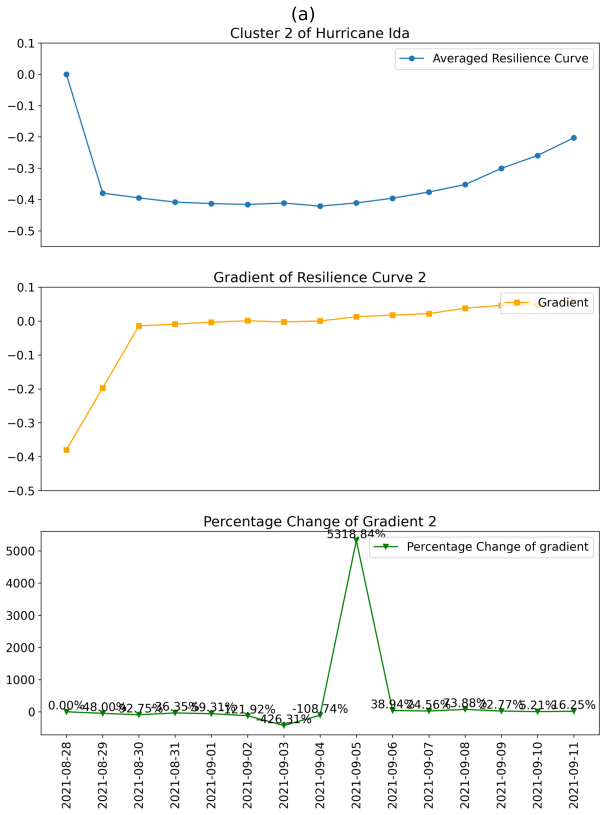
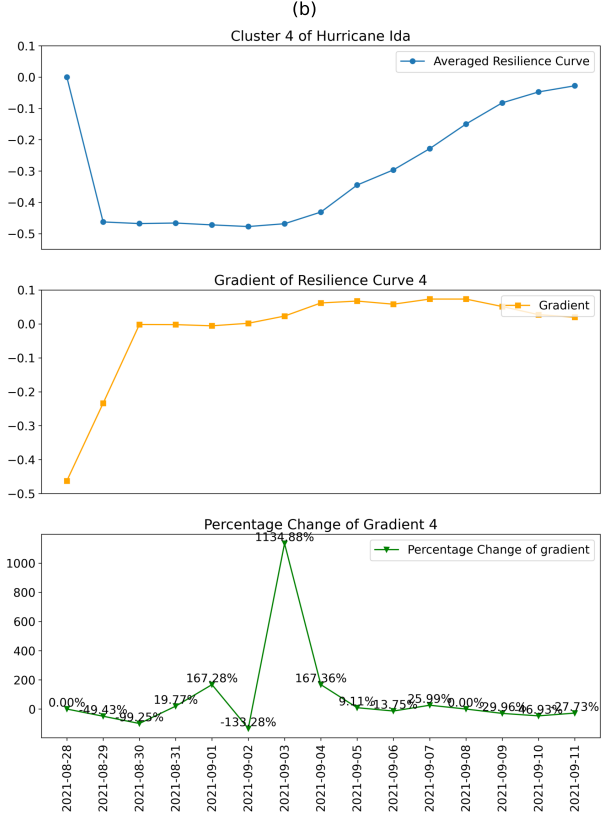
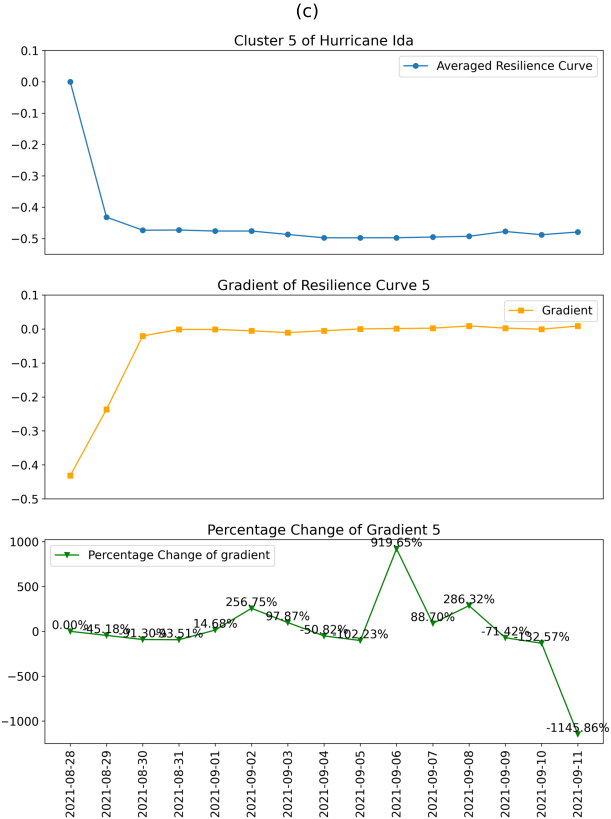

**Fig 8. Trapezoidal resilience curves and their properties.** Each subplot displays average resilience curve, gradients of each data point and gradient percentage changes of each data points. **a-c** display results for clusters 2, 4, and 5 in Hurricane Ida.

Similar analyses were performed on the triangular archetypes of power system resilience curves (cluster 1 in Hurricane Ida, clusters 1 and 2 in the ice storm, clusters 2 and 3 in Hurricane Irma), shown in Fig 9. Among the five curves, the gradient plots of four (cluster 1 in Hurricane Ida, cluster 1 in the ice storm, clusters 2 and 3 in Hurricane Irma, Fig.9a, 9c-9e) display common trends: after the gradients turn from negative to positive, the curves climb up initially and then decrease with a converging tendency to zero. The trajectory of gradient curves reflects a noteworthy system behavior that when the system starts to recover, the speed of recovery is not constant as with trapezoidal curves. Instead, the system performance represented by triangular curves first experiences rapid recovery until a certain performance level, and then the recovery rate slows down until full recovery. For example, between September 12 to September 15, 2021, cluster 2 of Hurricane Irma (Fig. 9d) recovers rapidly, with the average gradient value larger than 0.1 After September 15, the recovery gradients drop to around 0.01. Based on the same idea of separating sustained performance loss stage and recovery stage, we defined the turning point of recovery speed as recovery pivot, and the period before the point as rapid recovery stage and the other side as slow recovery stage. The corresponding performance level of recovery pivot point is -0.205 for cluster 2 and is -0.1980 for cluster 3 for Hurricane Irma. The results indicate that the critical functionality threshold was around 20% for the case of Hurricane Irma, and the power system would recover rapidly to the threshold to restore critical functions of the system, and then achieve full recovery with a slower pace. Clusters in the ice storm also follow the same pattern: cluster 1 has a recovery pivot point on February 4, 2023, at 18:00, with critical functionality threshold around 16.8%. For cluster 2 of the ice storm (Fig. 9b), no recovery pivot was identified, since the maximum power outage is 16.9%, which is almost equal to the critical functionality threshold. In other words, if the performance loss does not exceed the threshold, the power system would recover at a constant rate with no recovery pivot. The finding is consistent with the observation from both the average resilience curve and the gradient plot.

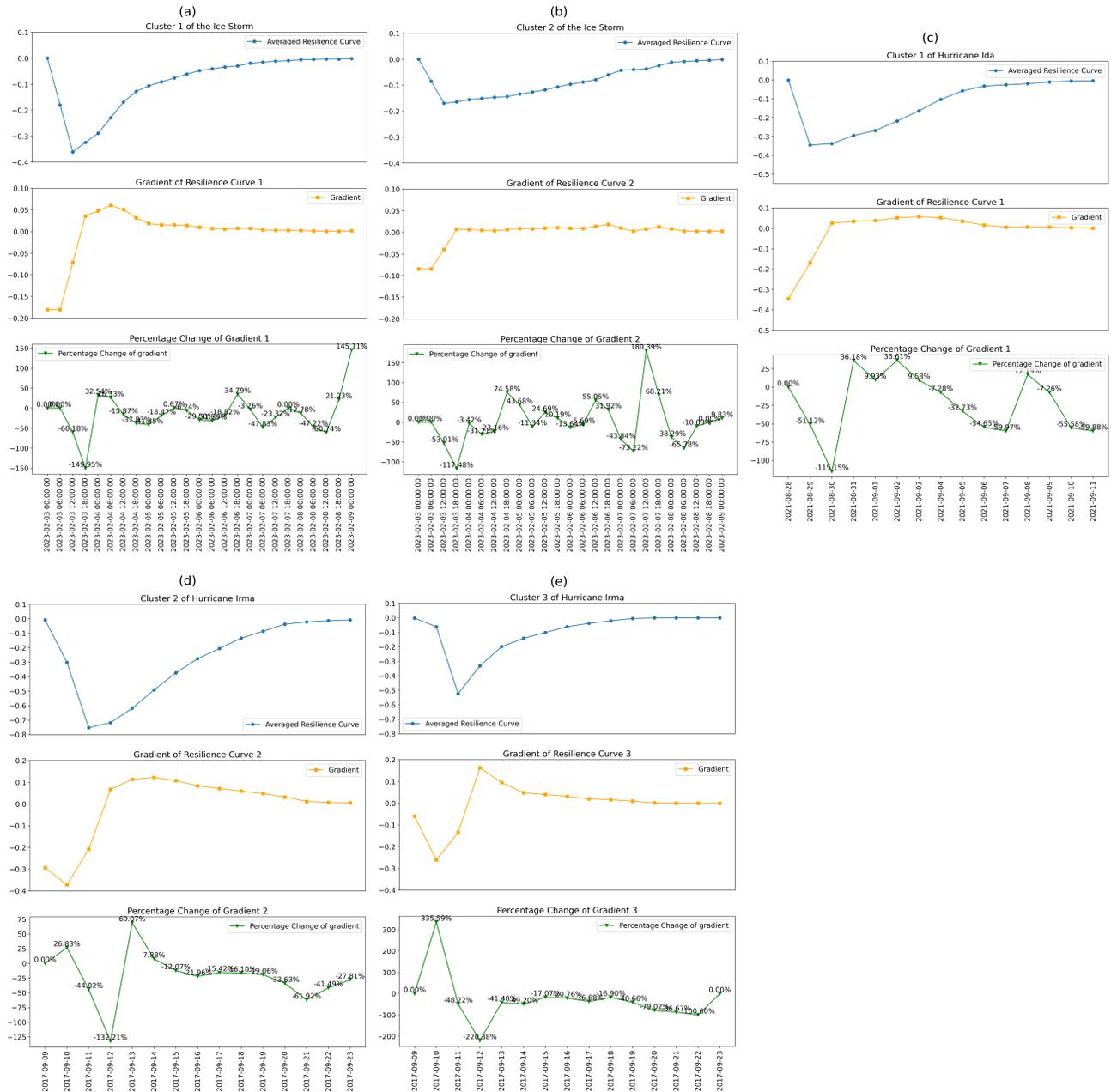

**Fig. 9 Triangular resilience curves and their properties.** Each subplot displays average resilience curve, gradients of each data point and gradient percentage changes of each data points. **a-e** display results for clusters 1 and 2 of the ice storm, cluster 1 of Hurricane Ida, and clusters 2 and 3 of Hurricane Irma.

Among all the clusters, cluster 1 of Hurricane Irma (Fig. 10) is a special case which we called a "transitional state" between triangular and trapezoidal curves, because it bears properties of both archetypes. From the resilience curve, it has a period with sustained performance loss, while the period is quite short, lasting only 6 hours. Two recovery pivots can be found on the curve, one separates the sustained performance loss stage and recovery stage, which is identical to trapezoidal curves. However, instead of having constant recovery rate, there exists another recovery pivot point, which separates rapid recovery with slow recovery; the corresponding threshold is around 12.9% power outage.

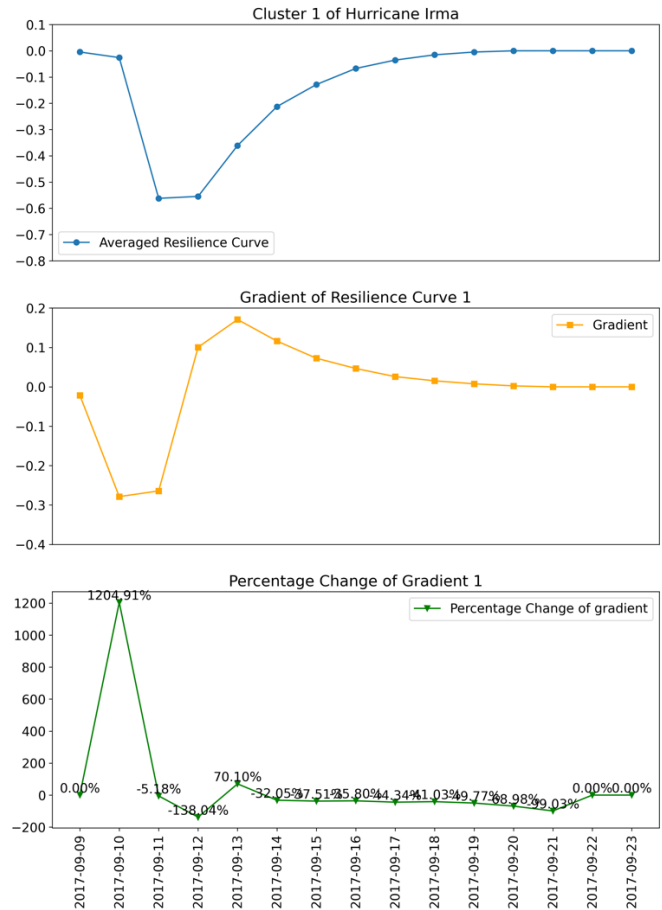

**Fig 10. Resilience curve and properties of cluster 1 during Hurricane Irma.**

## 5. Discussion and concluding remarks

Although resilience curves have remained the primary model for the understanding of resilience behavior of infrastructure systems, the most prevalent studies to characterize and quantify infrastructure resilience are proposed on a theoretical or simulation basis. Empirical research on resilience curve archetypes and their fundamental properties is rather limited. To fill the important gap, this study applied an unsupervised machine learning clustering method to examined more than 200 power system resilience curves related to power outages in three extreme weather events, which provides empirical support to facilitate the understanding of and prediction of resilience characteristics in infrastructure systems. We collected direct power outage data along with time stamps during Hurricane Ida, Hurricane Irma, and the ice storm across Texas in early 2023. Using power outage rate as an indicator of system performance, resilience curves of power systems at each geographical unit were delineated. Dynamic time warping based *k*-means, a method specifically applicable to time-series data, was applied to perform clustering. We examined the identified clusters to reveal resilience curve archetypes and evaluated their fundamental properties. The main findings of this study are threefold. First, this study identified two archetypes of power system resilience curves: triangular and trapezoidal curves, which verified the two theoretical models in the prior literature. Furthermore, this study also empirically found that the triangular curves and trapezoidal curves can be coincident within the same disruptive event. However, current analysis performed in this research could not specify factors that lead to the occurrence of triangular or trapezoidal resilience behaviors in different areas of a community. This limitation is because the power outage percentage is the only data we could obtain, and we could not collect data related to the intensity of the extreme weather event across the studied areas, the physical conditions of the systems, and the availability of resources after the event. If more data becomes available, future studies could examine factors that shape the occurrence of each resilience curve archetype.

Second, two fundamental properties determine the behavior of trapezoid curves: duration of sustained performance loss, and constant recovery rate. By direct observation and gradient, as well as by percentage change of gradient calculation, we found that the longer the sustained performance loss lasts, the slower the constant recovery rate would be. This finding suggest that a longer period of sustained performance loss could be an indicator of a greater extent of damage which would lead to a slower recovery rate. Similarly, three fundamental properties affecting resilience behaviors exist for triangular curves: recovery pivot point, critical functionality threshold, and critical functionality recovery rate. Similar to trapezoidal curves, the recovery pivot point denotes the time point when the recovery rate changes. Remarkably, the recovery pivot point occurs when the system reaches the critical functionality threshold, which is about 80% to 90% (namely 10% to 20% performance loss) in this case. When about 80% to 90% of the system performance is restored, the recovery rate slows done, and hence recovery rate changes. If the actual performance loss exceeds the critical functionality threshold, the recovery would proceed in a rapid manner until it reaches the recovery pivot point. For the cases studied in this research, the value of the critical functionality threshold is between 80% to 90%. However, the universality of this critical functionality threshold for power infrastructure needs to be verified using additional datasets related to power outages. If the actual performance does not drop below the critical threshold, the system will recover quickly with a constant rate. If the performance loss is greater than the critical threshold, the recovery would follow a bi-modal recovery rate with the recovery pivot point occurring when the system reaches the critical threshold. The results of this research deepen our understanding of the resilience performance of power infrastructure systems and provide a more detailed characterization of infrastructure resilience curves beyond a mere conceptual visual representation. By identifying the fundamental resilience curve archetypes and their associated properties, the findings provide fresh and novel insights for researchers and practitioners to characterize and predict infrastructure resilience performance. The methods for characterizing resilience curve archetypes used in this study could be used in future studies in studying resilience behaviors in other infrastructure systems. Such characterizations would move us closer to a deeper and more detailed understanding of the resilience behavior of infrastructure during disruptive events.

Finally, the presented study has several limitations that could be addressed in future research. For example, the study is limited by the resolution and scale of datasets. Power outage data is highly perishable and difficult to obtain at a fine resolution. All the power outage data were collected manually by the authors, which is not only time consuming, but also limits the spatial resolution of the data provided by utilities on their websites. If power outage datasets with a greater spatial resolution are available, future studies can build upon the findings of this study and perhaps explore additional key characteristics in the resilience behaviors of power infrastructure. Also, in the absence of additional data regarding the power infrastructure operators' restoration strategies and damage levels to different subsystems, we could not specify the factor that led to the occurrence of each resilience curve archetype (triangular versus trapezoidal). If possible, future studies could gather such information to uncover the factors that shape which resilience curve archetype is manifested.

**Acknowledgement**

This material is based in part upon work supported by the National Science Foundation under Grant CMMI-1846069 (CAREER). Any opinions, findings, conclusions, or recommendations expressed in this material are those of the authors and do not necessarily reflect the views of the National Science Foundation.